\documentclass[11pt,a4paper]{article}
\usepackage[hyperref]{acl2021}
\usepackage{times}
\usepackage{latexsym}

\usepackage{amsmath,amsthm,amssymb,amsfonts}
\usepackage{booktabs}
\usepackage{multirow}
\usepackage{graphicx}

\usepackage{makecell}
\usepackage{enumitem}
\usepackage{color}
\usepackage{comment}
\usepackage{subcaption}
\usepackage{wrapfig}

\aclfinalcopy

\title{Sentence Semantic Regression for Text Generation}

\author{Wei Wang$^{1,}$\footnotemark[1]\thanks{~~Work was done during internship at Tencent AI Lab.}~, Piji Li$^2$, Hai-Tao Zheng$^1$ \\
$^1$Shenzhen International Graduate School, Tsinghua University \\ 
$^2$Tencent AI Lab \\
 {\tt w-w16@mails.tsinghua.edu.cn} \\
  {\tt lipiji.pz@gmail.com} \\
   {\tt zheng.haitao@sz.tsinghua.edu.cn}
}

\begin{document}
\maketitle
\begin{abstract}

Recall the classical text generation works, the generation framework can be briefly divided into two phases: \textbf{idea reasoning} and \textbf{surface realization}. The target of idea reasoning is to figure out the main idea which will be presented in the following talking/writing periods. Surface realization aims to arrange the most appropriate sentence to depict and convey the information distilled from the main idea. However, the current popular token-by-token text generation methods ignore this crucial process and suffer from many serious issues, such as idea/topic drift. To tackle the problems and realize this two-phase paradigm, we propose a new framework named Sentence Semantic Regression (\textbf{SSR}) based on sentence-level language modeling. For idea reasoning, two architectures \textbf{SSR-AR} and \textbf{SSR-NonAR} are designed to conduct sentence semantic regression autoregressively (like GPT2/3) and bidirectionally (like BERT). In the phase of surface realization, a mixed-granularity sentence decoder is designed to generate text with better consistency by jointly incorporating the predicted sentence-level main idea as well as the preceding contextual token-level information. We conduct experiments on four tasks of story ending prediction, story ending generation, dialogue generation, and sentence infilling. The results show that SSR can obtain better performance in terms of automatic metrics and human evaluation.
\end{abstract}

\section{Introduction}
\label{sec:intro}

The past few years have seen tremendous progress in the area of text generation, such as dialogue generation~\citep{shang2015neural,wang2020large,zhang2020dialogpt,roller2020recipes,adiwardana2020towards}, abstractive summarization~\citep{rush2015neural,zhang2020pegasus}, story generation~\citep{chaturvedi2017story,fan2018hierarchical,zhou2019story,see2019massively,tan2020progressive}, image/video captioning~\citep{vinyals2016show,you2016image}, neural machine translation~\citep{DBLP:journals/corr/BahdanauCB14,vaswani2017attention,gu2017non}, etc. A large number of researchers have been attracted to this area to fix the potential problems and improve the performance. More recently, the large generative pre-trained language models, such as GPT2/3~\citep{radford2019language,brown2020language}, have greatly improved the quality of text generation and can be used to solve different text generation tasks simultaneously in a unified framework.

\begin{figure}[t!]
\centering
\includegraphics[width=0.8\columnwidth]{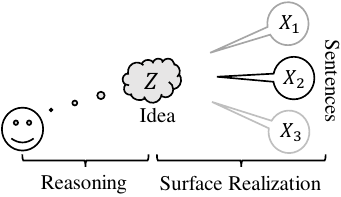}
\caption{Text producing process.}
\label{fig:front}
\vspace{-8mm}
\end{figure}

Let us recall those pioneer classical text generation works~\citep{mann1983overview,mckeown1992text,reiter1997building,gatt2018survey}, where, generally and briefly, the process of text generation can be divided into two phases, as depicted in Figure~\ref{fig:front}: \textbf{idea reasoning} and \textbf{surface realization}. The target of ``idea reasoning'' is to figure out the main idea/topic which will be presented in the following talking/writing periods. The main idea can be independent, or dependent on the context such as the last conversation utterance. After careful consideration, we obtain the main idea $Z$. Then, the objective of the second phase is to arrange the most appropriate surface symbol sequence to depict and convey the information distilled from the main idea, which is named ``surface realization''. Usually, the surface symbol sequence could be one kind of natural languages, e.g., English.

More importantly, the results of surface realization for the same idea $Z$ can be different. We may jointly consider many factors to select the most appropriate result for the current scenario. The factors could be relevance, discourse consistency, coreference, ethical issues, etc.
After those two phases, the best sentence used to present the main idea is produced. 
Indeed, the above-mentioned two-phase framework is a natural and crucial manner to conduct the problem of text generation. Therefore, in this work, we dedicate to investigating the technical solutions to realize the two-phase text generation framework.

Let us rethink the current popular text generation techniques. However, during the investigations, we find that most of the text generation approaches, such as the pre-trained auto-regressive language models GPT2/3~\citep{radford2019language,brown2020language} as well as the typical decoders used in the seq2seq frameworks, are all trained in the paradigm of token-level language model (here, token can be BPE~\citep{sennrich2016neural} token). During the inference stage, sentences are generated in a manner of token-by-token, autoregressively. Even though it has been proved that this paradigm can generate convincing text with high fluency and diversity, they are still suffering from many serious issues such as repetition~\citep{holtzman2019curious,welleck2019neural,fu2020theoretical}, hallucinations~\citep{nie-etal-2019-simple,maynez2020faithfulness}, illogical issues~\citep{floridi2020gpt}, etc. 
Therefore, we claim that token-level generation models cannot capture the high-level semantic information between sentences. And in surface realization, the optimization objective at the token level focuses more on the neighbor token consistency instead of idea/topic consistency. Moreover, for some open-ended text generation tasks such as story generation~\citep{fan2018hierarchical,see2019massively}, due to the sampling-based decoding strategy (such as top-k~\citep{fan2018hierarchical} and top-p~\citep{holtzman2019curious}), idea/topic drift easily occurs, then the generated sentence cannot depict the main idea/topic faithfully.
Actually, for text producing, humans will get the main idea first (\textbf{idea reasoning}), and then translate it using a token sequence (\textbf{surface realization}), rather than produce tokens one-by-one by looking back at the previous tokens.

To tackle the above-mentioned problem, to realize the two-phase text generation framework, we explore and propose a new generation paradigm called Sentence Semantic Regression (\textbf{SSR}). Our approach consists of three components: a sentence encoder, a sentence-level language model (for \textbf{idea reasoning}), and a mixed-granularity sentence decoder (for \textbf{surface realization}). In the idea reasoning phase, the sentence encoder is used to convert the sentence into a semantic vector. And the sentence-level language model conducts the \textit{sentence semantic regression} on the obtained sentence vector sequence and then predicts the target sentence vector which is regarded as the main idea to be described. In the surface realization phase, the mixed-granularity decoder generates the sentence token by token mainly conditioning on the predicted sentence vector. In order to introduce token-level consistency between sentences, tokens from the previous sentences are also used as the context of the decoder.
Inspired by the token-level autoregressive language models (GPT2/3~\citep{radford2019language,brown2020language}) and bidirectional non-autoregressive masked language models (BERT~\citep{devlin2019bert}), we also design two sentence-level language models, named \textbf{SSR-AR} (like GPT2/3) and \textbf{SSR-NonAR} (like BERT), to conduct idea reasoning autoregressively or bidirectionally. We conduct experiments on the tasks of story ending prediction, story ending generation, dialogue generation, and sentence infilling. The results show that our proposed framework SSR can obtain better performance in terms of automatic metrics and human evaluation.

In summary, our contributions are as follows:
\begin{itemize}[leftmargin=*,topsep=0pt]
    \setlength\itemsep{-0.35em}
    \item Text generation generally consists of two phases: idea reasoning and surface realization. To realize this two-phase paradigm, we propose a new framework named Sentence Semantic Regression (\textbf{SSR}) based on sentence-level language modeling.
    \item For idea reasoning, two architectures \textbf{SSR-AR} and \textbf{SSR-NonAR} are designed to conduct sentence semantic regression autoregressively and bidirectionally.
    \item For surface realization, a mixed-granularity sentence decoder is designed to generate text with better consistency by jointly incorporating the predicted sentence-level main idea as well as the tokens of the previous context.
    \item Experimental results on four different tasks demonstrate the effectiveness of our framework.
\end{itemize}


\begin{figure*}[t!]
\centering
\includegraphics[width=1.8\columnwidth]{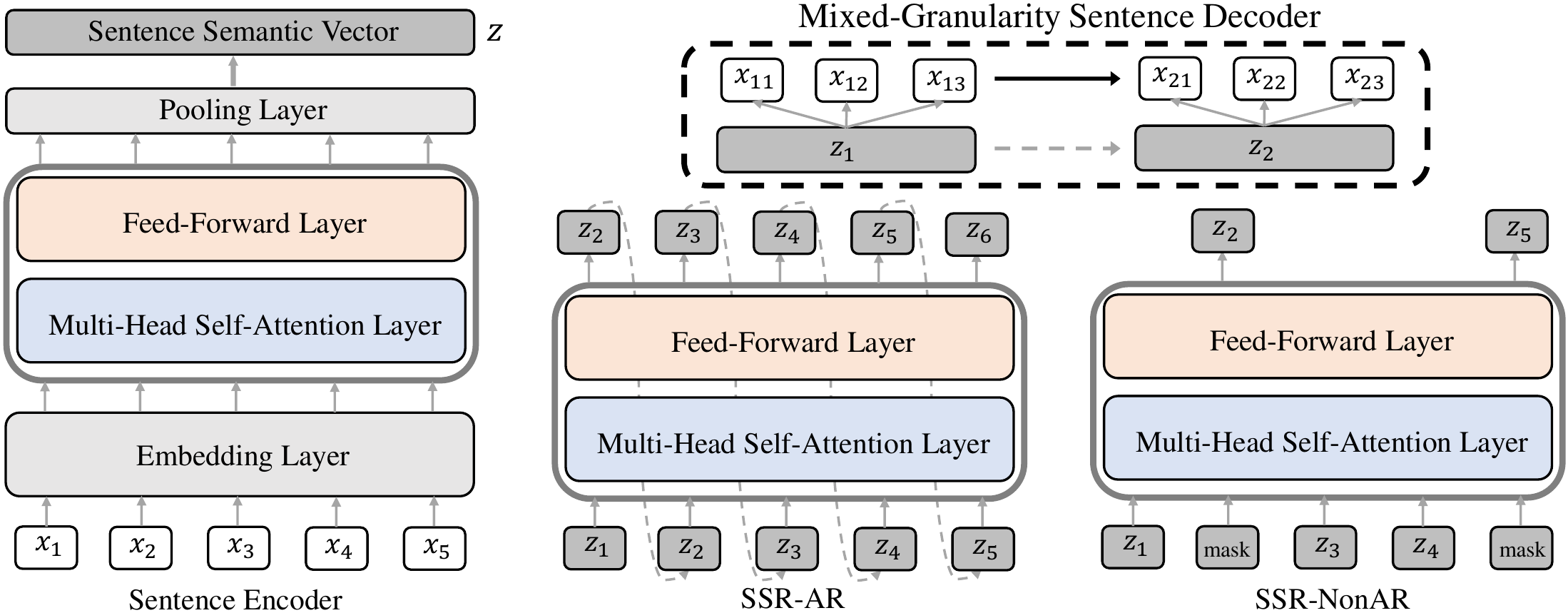}
\caption{Illustration of our proposed framework SSR: Sentence Encoder, SSR-AR, SSR-NonAR, and Mixed-Granularity Sentence Decoder.}
\label{fig:framework}
\vspace{-5mm}
\end{figure*}

\section{The Proposed SSR Framework}

\subsection{Overview}

We denote a sentence as $X = (x_1, x_2, \ldots, x_T)$, where $x_i$ represents the $i$th token, $T$ is the sentence length. A set of sentences is denoted as $\mathcal{X} = (X_1, X_2, \ldots, X_N)$. Our goal is to generate the last sentence $X_N$ given the previous $N-1$ sentences (next sentence generation) or to generate the sentence $X_i$ given the other surrounding sentences (sentence infilling).
As shown in Figure~\ref{fig:framework}, our \textbf{S}entence \textbf{S}emantic \textbf{R}egression (\textbf{SSR}) framework consists of three components: sentence encoder, sentence-level language model, and mixed-granularity sentence decoder. We first use the sentence encoder to convert sentences into corresponding sentence vectors. Then the sentence-level language model conducts language modeling on sentence vector sequences and learns to predict sentence vectors to represent semantic information about the text to be generated. Finally, the mixed-granularity sentence decoder generates a sentence token by token conditioning on the predicted sentence vector. We consider two sentence-level language models, called SSR-AR and SSR-NonAR, to utilize two types of objectives of language modeling, autoregressive modeling \citep{radford2019language,brown2020language} and non-autoregressive masked modeling \citep{devlin2019bert}. The SSR-AR is mainly used for next sentence generation task and the SSR-NonAR is used for sentence infilling.

\subsection{Transformer-based Language Models}

Recently, autoregressive language models such as GPT2/3~\citep{radford2019language,brown2020language} and non-autoregressive bidirectional language models such as BERT~\citep{devlin2019bert} are the most typical paradigms designed for language generation and understanding, respectively. The backbones of those two LM branches are based on Transformer~\citep{vaswani2017attention}.

Transformer consists of $N$ identical self-attention blocks and each block contains two sub-layers: a multi-head attention layer and a feed-forward layer. An add \& norm layer is employed around each of those two sub-layers. Formally, given the input vectors from the previous block $\mathbf{H}^{n-1}$, the output $\mathbf{H}^{n}$ of the current block is computed as follows:
\begin{align}
\setlength{\abovedisplayskip}{3pt}
\setlength{\belowdisplayskip}{3pt}
\small
	\mathbf{C}^{n} & = \operatorname{LN}\left(\operatorname{SELF-ATT}\left(\mathbf{H}^{n-1}\right)+\mathbf{H}^{n-1}\right) \\
	\mathbf{H}^{n} &=\operatorname{LN}\left(\operatorname{FFN}\left(\mathbf{C}^{n}\right)+\mathbf{C}^{n}\right)
\end{align}
where  $\operatorname{SELF-ATT}(\cdot)$,  $\operatorname{LN}(\cdot)$,  and  $\operatorname{FFN}(\cdot)$ are  respectively  self-attention  mechanism,  layer normalization, and feed-forward network.  $\operatorname{SELF-ATT}(\cdot)$ conducts attention modeling over the input $\mathbf{H}^{n-1}$:
\begin{equation}
\setlength{\abovedisplayskip}{3pt}
\setlength{\belowdisplayskip}{3pt}
\small
    \operatorname{SELF-ATT}\left(\mathbf{H}^{n-1}\right)=\operatorname{softmax}\left(\frac{\mathbf{Q} \mathbf{K}^{\top}}{\sqrt{d_{k}}}\right) \mathbf{V}
\end{equation}
where $\{\mathbf{Q,K,V}\}$ are query, key, and value vectors which are transformed from the input variable $\mathbf{H}^{n-1}$. $\sqrt{d_k}$ is the scaling factor where the $d_k$ is the dimension size of the query and key vectors. There are two learnable embedding tables, word embedding table and positional embedding table. Given the word embeddings $\mathbf{E}=\{\mathbf{e}_1,\mathbf{e}_2,...,\mathbf{e}_T\}$ and corresponding positional embeddings $\mathbf{P}=\{\mathbf{p}_1, \mathbf{p}_2,...,\mathbf{p}_T\}$ of a sentence, the input for the first block is $\mathbf{H}^0=\mathbf{E}+ \mathbf{P}$. 

Finally, a linear function $g$ with $\operatorname{softmax}$ activation is used to compute the probability of next token $x_t$ for GPT2/3 via:
\begin{equation}
\setlength{\abovedisplayskip}{3pt}
\setlength{\belowdisplayskip}{3pt}
\small
\label{eq:outprob}
	p \left( x_t | x_{\leq{t-1}}  \right) = \operatorname { softmax } \left( g \left( \mathbf{h}_ { t } \right) \right)
\end{equation}
or predict the masked token $x_t$ for BERT via:
\begin{equation}
\setlength{\abovedisplayskip}{3pt}
\setlength{\belowdisplayskip}{3pt}
\small
\label{eq:outprob_bert}
	p \left( x_t | x_{\leq{t-1}},x_{\geq{t+1}}  \right) = \operatorname { softmax } \left( g \left( \mathbf{h}_ { t } \right) \right)
\end{equation}

We utilize the negative log-likelihood (NLL) loss as the optimization objective to train the model:
\begin{equation}
\setlength{\abovedisplayskip}{3pt}
\setlength{\belowdisplayskip}{3pt}
\small
	\mathcal{L}_{\mathrm{lm}}=- \frac{1}{T}\sum_t \log p\left(x_t\right)
	\label{eq:nll_loss}
\end{equation}
where $p(x_t)$ can be $p( x_t | x_{\leq{t-1}})$ for GPT2/3 or $p ( x_t | x_{\leq{t-1}},x_{\geq{t+1}} )$ for BERT.

\subsection{Sentence Semantic Encoder}\label{sec:encoder}
Our framework treats a sentence as a semantic unit and conducts language modeling on sentence level instead of token level. Sentences are converted into dense vectors to represent the semantic information. The quality of sentence representation directly affects the performance of sentence-level language modeling. In this work, we use BERT to obtain sentence vectors, considering that BERT is good at understanding and is widely used in many tasks. Certainly, we could use other sentence representation methods \citep{reimers-gurevych-2019-sentence,zhang2020unsupervised,li2020sentence, gao2021simcse}. Considering that it is not the main focus of this work, we leave them for future investigations.

Specifically, as shown in Figure~\ref{fig:framework}, we first load the pre-trained weights of BERT and make the parameter fixed. Then we feed a sentence into BERT and obtain hidden states of the second-to-last layer for each word as $\mathbf{H}^{L-1}$. We apply mean-pooling on word vectors $\mathbf{H}^{L-1}$ to obtain the sentence vector $\mathbf{z}_i$ for the $i$th sentence. 

\subsection{Sentence Semantic Regression (SSR)}
The core idea of Sentence Semantic Regression is sentence-level language modeling. The sentence-level language model conducts language modeling on sentence vector sequences and learns to predict the target sentence vector as the main idea to be presented and conveyed. As shown in Figure~\ref{fig:framework}, we consider two types of objectives for language modeling, autoregressive modeling \citep{radford2019language,brown2020language} and non-autoregressive masked modeling \citep{devlin2019bert}. The former is called \textbf{SSR-AR} and the latter is called \textbf{SSR-NonAR}. SSR-AR applies an architecture similar to GPT2/3 and SSR-NonAR is similar to BERT. There are several differences between our sentence-level language models and GPT2/3 or BERT. 

First, after obtaining the sentence semantic vectors $\mathbf{Z}$, we directly feed them to the model of SSR and our model has no word embedding layer. The positional embedding table still remains to indicate sentence location in the original paragraph. Specifically, given the sentence vectors $\mathbf{Z}=\{\mathbf{z}_1,\mathbf{z}_2,...,\mathbf{z}_m\}$ and corresponding positional embeddings $\mathbf{P^s}=\{\mathbf{p}^s_1,\mathbf{p}^s_2,...,\mathbf{p}^s_m\}$ of a paragraph with $m$ sentences, the inputs of the first block are $\mathbf{H}^0=\mathbf{Z}+ \mathbf{P^s}$.

Second, different with GPT/BERT, we remove the \texttt{softmax} activation function in the output layer and predict a vector with the original dimension size $d$ as follows:
\begin{equation}
\setlength{\abovedisplayskip}{3pt}
\setlength{\belowdisplayskip}{3pt}
\small
    \tilde{\mathbf{z}}_t = g'(\mathbf{h}^s_t)
\end{equation}

Third, the loss function is also different with GPT/BERT considering that the target of SSR is to predict a vector $ \tilde{\mathbf{z}}_t$. We employ cosine similarity between the predicted sentence vector $\tilde{\mathbf{z}}_t$ and the ground truth vector $\mathbf{z}_t$ as the optimization objective, and the calculation is as follows:
\begin{equation}
\setlength{\abovedisplayskip}{3pt}
\setlength{\belowdisplayskip}{3pt}
\small
    \mathcal{L}_{\mathrm{ssr}}= \frac{1}{m}\sum_t (1- \operatorname{cos}(\tilde{\mathbf{z}}_t, \mathbf{z}_t))
    \label{eq:loss_cosine}
\end{equation}
And this is the main reason we name our framework as ``\textbf{Sentence Semantic Regression}''.

Generally, in GPT2/3 and BERT, the cross entropy loss optimizes the positive tokens and the negative tokens simultaneously because of the superior property of \texttt{softmax} operation. However, our cosine similarity-based loss only takes into account the positive sentences, ignoring the negative ones. Thus, to enhance the performance, inspired by the contrastive learning paradigm~\citep{he2020momentum,chen2020simple}, we randomly sample $n$ sentences from other paragraphs for each sample as negative sentences. Then the contrastive loss function with negative examples is changed to:
\begin{equation}
\setlength{\abovedisplayskip}{3pt}
\setlength{\belowdisplayskip}{3pt}
\small
    \mathcal{L}^{contrast}_{\mathrm{ssr}} = 
    \frac{1}{m}\sum_t \left(2- \operatorname{cos}(\tilde{\mathbf{z}}_t, \mathbf{z}_t) + \frac{1}{n}\sum_i \operatorname{cos}(\tilde{\mathbf{z}}_t, \mathbf{z}_i)\right)
    \label{eq:negative_loss}
\end{equation}
where $\tilde{\mathbf{z}}_t$ is the predicted vector, $\mathbf{z}_t$ is the ground truth vector, and $\mathbf{z}_i$ is the negative sample.

\textbf{SSR-AR}
Similar to the autoregressive language models such as GPT2/3, we conduct sentence semantic regression autoregressively. Here, the input is a sequence of sentence semantic vectors, and the target is the left-shifting of the input.

\textbf{SSR-NonAR}
As BERT, for the input sentence vector sequence, we randomly mask off 15\% of them with zero vectors and then use the rest of the sentences to predict the masked sentence vectors.

\subsection{Surface Realization}
For our framework SSR, surface realization can be implemented via two kinds of approaches: \textbf{matching} and \textbf{generation}.

\textbf{Surface Realization via Matching}
Since we have obtained the predicted sentence vector $\tilde{\mathbf{z}}$ (the main idea to be presented), we can find a sentence $X'$ from a candidate corpus whose semantic vector $\mathbf{z}'$ is the most matched one with $\tilde{\mathbf{z}}$. The matching algorithm can be cosine similarity, inner product, or even a trained matching model. Here, we claim that matching is just a possible approach to find a candidate sentence (we designed experiments to clarify this point), and certainly it is not the most appropriate one due to some issues such as coreference and discourse consistency.

\textbf{Surface Realization via Generation}
We believe generation is a better manner to conduct surface realization.
Thus, we build a Mixed-Granularity (sentence- and token-granularity) sentence decoder, which is used to recover the corresponding sentence token by token conditioning on the predicted sentence vector by SSR-AR or SSR-NonAR. The backbone of the decoder is still a Transformer-based autoregressive language model.

Specifically, we use the predicted sentence vector $\tilde{\mathbf{z}}$ as the first token and \texttt{<eos>} as the second token, which are regarded as context information and fed into the decoder to conduct generation:
\begin{equation}
\setlength{\abovedisplayskip}{3pt}
\setlength{\belowdisplayskip}{3pt}
    \tilde{\mathbf{z}}, \texttt{<eos>} \longrightarrow y_{1}, y_{2},\ldots,y_{T'}, \texttt{<eos>}
    \label{eq:vanilla_decoder}
\end{equation}
We call this version a vanilla sentence decoder. Although the input format is similar to GPT2, there is still a big difference between the two methods. Here, our vanilla sentence decoder only conducts generation according to the sentence vector $\mathbf{z}$ and restores the semantic information as much as possible. However, we found that the decoder could not accurately restore some detailed low-frequency information, such as the named entities which usually appear in the previous sentences. Moreover, as mentioned in Section~\ref{sec:intro}, the results of surface realization for the same idea $\mathbf{z}$ can be different. Therefore, in order to generate an appropriate result for the current context, some factors such as token-level discourse/coreference consistency need to be recalled.

\textbf{Consistency Enhanced Realization}
Thus, we enrich the context by briefly appending the tokens from the previous sentences before \texttt{<eos>} to conduct generation:
\begin{align}
\setlength{\abovedisplayskip}{3pt}
\setlength{\belowdisplayskip}{3pt}
\begin{split}
    \tilde{\mathbf{z}},&..., x_{l-1,n-1}, x_{l-1,n}, \texttt{<eos>} \\
    &\longrightarrow y_{l,1}, y_{l,2},...,y_{l,T'},\texttt{<eos>}
\end{split}
\label{eq:mixed_decoder}
\end{align}

Since both the sentence-level information ($\tilde{\mathbf{z}}$) and token-level information are used as context to conduct generation, we call it Mixed-Granularity Sentence Decoder. In experiments, \textit{we interestingly find that the mixed-granularity sentence decoder can restore named entities accurately}.

\subsection{Training}
The sentence encoder loads the pre-trained BERT directly and we do not fine-tune it. We first reprocess the data with the sentence encoder to get all the sentence vectors. Then we use sentence vector sequences to train the sentence-level language models SSR-AR and SSR-NonAR, and regrade $\mathcal{L}^{contrast}_{\mathrm{ssr}}$ as the loss function. For the decoder, (the sentence vector of target sentence, the tokens of previous sentences, the tokens of target sentence) triples are used to train the mixed-granularity sentence decoder and the typical NLL loss is regarded as the optimization objective.

\section{Experimental Setup}

In order to evaluate our framework, we conduct experiments on four different tasks, story ending prediction, story ending generation, dialogue generation, and sentence infilling. Story ending prediction, also called Story Cloze, aims to select the coherent ending from two candidate sentences. This task is used to verify the capability of SSR with \textit{surface realization via matching}.
Story ending generation and dialogue generation belong to next sentence generation, which aims to generate a final sentence given several previous sentences. We use these two tasks to evaluate SSR-AR. Sentence infilling aims to generate an intermediate sentence given the surrounding sentences, which goal is in line with SSR-NonAR's objective. Therefore, we use this task to validate SSR-NonAR.


\subsection{Datasets}\label{sec:datasets}

For story ending prediction, story ending generation, and sentence infilling, we use the ROC Stories dataset, which consists of stories focusing on common sense \citep{mostafazadeh2016corpus}.
The dataset was introduced for the Story Cloze task. 
The training set has 98k stories and each story has five sentences. The validation set 2016, test set 2016, and validation set 2018 contain 1.8k/1.8k/1.5k stories consisting of four sentences followed by two alternative endings: one ending is coherent with the context; the other is not. 
Notice that there are no labels in the training set.

For story ending prediction, we follow the original task objective.
We conduct unsupervised training on the training set and then evaluate the model on validation and test sets.
Many works on this task used a supervised setting and they used the validation set as a small labeled training set to train the model \citep{chaturvedi2017story,zhou2019story,li2019story, cui2020discriminative}. Therefore, we don't compare with these methods. For story ending generation and sentence infilling, we split the original training set into TRAIN, VAL, and TEST (88k/5k/5k).
For dialogue generation, we use the Dailydialog dataset \citep{li2017dailydialog}. Dailydialog contains 13k daily conversations under ten different topics. Statistics show that the speaker turns are roughly 8, and the average number of tokens per utterance is about 15. The dataset is randomly separated into TRAIN, VAL, and TEST (11k/1k/1k).

In order to evaluate that our framework can improve the performance by pre-training the sentence-level language model, we use the BooksCorpus ~\citep{zhu2015aligning} and English Wikipedia\footnote{These two datasets are available in huggingface/datasets} to pre-train SSR-NonAR.
Specifically, we split the text of the two corpora into sentences, and then use the sentence encoder to obtain all the sentence vectors. We use 128 consecutive sentences as a paragraph to train SSR-NonAR.

\subsection{Baseline Models}

For four tasks, we mainly compare our framework with a common token-level language model, \textbf{GPT2} \citep{radford2019language}. On the one hand, GPT2 is better than many task-specific models for many generation tasks. On the other hand, we mainly want to prove that the sentence-level language model is superior to the token-level language model in text generation tasks. Therefore, we do not compare with some other methods. It should be noted that GPT2 cannot be used directly for sentence infilling, while IGPT2 proposed by \citet{donahue-etal-2020-enabling} designs a mask strategy so that it can be used for filling blanks with GPT2. Therefore, we compare our framework with \textbf{IGPT2} in sentence infilling task. \textbf{GPT2(FT)} and \textbf{IGPT2(FT)} mean that we fine-tune the model on the corresponding task data. 

In order to evaluate each component of SSR, we conduct a detailed ablation analysis for the following variants. \textbf{SSR(I)} denotes the basic version of our framework, which only uses the cosine loss $\mathcal{L}_{\mathrm{ssr}}$ (Eq.(\ref{eq:loss_cosine})) and the vanilla sentence decoder (Eq.(\ref{eq:vanilla_decoder})). \textbf{SSR(II)} denotes our framework with the contrastive loss $\mathcal{L}^{contrast}_{\mathrm{ssr}}$ (Eq.(\ref{eq:negative_loss})) and the vanilla decoder. \textbf{SSR} is our full framework, which utilizes the contrastive loss and the mixed-granularity sentence decoder (Eq.(\ref{eq:mixed_decoder})). \textbf{PT} means that we first pre-train our SSR on BooksCorpus and Wikipedia.

\begin{table*}[!t]
\renewcommand\arraystretch{0.8}
\centering
\small
\resizebox{1.8\columnwidth}{!}{
\begin{tabular}{@{}l|l|ccc|c@{}}
\Xhline{3\arrayrulewidth}
\textbf{Type} & \textbf{Model}  & \textbf{Valid2016} & \textbf{Test2016} & \textbf{Valid2018} & \textbf{Avg}  \\
                                        \hline
\multirow{3}{*}{Token-Level LMs} & Schwartz et al. (2017)       &  -          & 67.7      &   -         & -     \\
                                        & GPT2  & 57.4       & 59.2      & 57.7       & 58.1 \\
                                        & GPT2(FT)           & 68.5       & 69.1      & 69.1       & 68.9 \\  \hline
SSR-AR                               &  SSR-AR(I)                            & 67.1       & 67.0      & 67.7       & 67.3 \\
                                        & SSR-AR(II)             & \underline{72.7}       & 71.0      & \underline{72.4}       & \underline{72.0} \\
                                         \hline
SSR-NonAR                                &  SSR-NonAR(I)                            & 66.4       & 66.9      & 66.9       & 66.7 \\
                                        & SSR-NonAR(II)             & 72.0       & \underline{71.7}      & 72.1       & \underline{72.0} \\
                                        & SSR-NonAR(II)+PT                    & \textbf{73.7}       & \textbf{72.1}      & \textbf{73.3}       & \textbf{73.0} \\
\Xhline{3\arrayrulewidth}
\end{tabular}
}
\vspace{-1mm}
\caption{Automatic evaluation results of story ending prediction.}
\label{table:story_ending_prediction}
\vspace{-3mm}
\end{table*}

\begin{table*}[!t]
\renewcommand\arraystretch{0.8}
\centering
\small
\resizebox{1.8\columnwidth}{!}{
\begin{tabular}{@{}l|cccc|cccc@{}}
\Xhline{3\arrayrulewidth}
                  & \textbf{B-1} &\textbf{B-2} &\textbf{B-3} & \textbf{B-4} & \textbf{D-1} & \textbf{D-2}  & \textbf{D-3}  & \textbf{D-4}  \\
\hline
GPT2(FT)  & 0.1180 & \underline{0.0223} & \underline{0.0053} & \textbf{0.0015} & 0.1379     & 0.5185     & 0.8329     & 0.9546     \\
\hline
SSR-AR(I)         & \underline{0.1265} & 0.0163 & 0.0026 & 0.0005 & 0.0802     & 0.3297     & 0.6170     & 0.8036     \\
SSR-AR(II) & 0.1018 & 0.0150 & 0.0022 & 0.0005 & \underline{0.1512}     & \underline{0.5425}     & \underline{0.8603}     & \underline{0.9665}     \\
SSR-AR  & \textbf{0.1281} &\textbf{ 0.0271} & \textbf{0.0061} & \underline{0.0014} & \textbf{0.1617}     & \textbf{0.5650}     & \textbf{0.8728}     & \textbf{0.9709}     \\
\Xhline{3\arrayrulewidth}
\end{tabular}
}
\vspace{-1mm}
\caption{Automatic evaluation results of story ending generation.}
\label{table:story_ending_generaton}
\vspace{-6mm}
\end{table*}

\subsection{Evaluation Metrics and Implementation Details}\label{sec:human_eval}\label{sec:exp_detail}

\noindent\textbf{Automatic Evaluation.}
For story ending prediction, we evaluate the accuracy of binary classification. For story ending generation, dialogue generation, and sentence infilling, we use BLEU \citep{papineni2002bleu} and Distinct \citep{li2016persona} as metrics.
BLEU measures the n-gram overlap between generated text and ground truth.
Distinct aims to evaluate the diversity of generated text.

\noindent\textbf{Human Evaluation.}
To further evaluate the quality of generated text in three generation tasks, we conduct pair-wise comparisons by human evaluation. We evaluate the models from the following three perspectives: \textbf{Grammaticality} to indicate whether a story is natural and fluent, \textbf{Logicality} to indicate whether a story is consistent and coherent in terms of causal dependencies in the context, and \textbf{Informativeness} to evaluate how much novel information the generated text contains. Note that the three aspects are independently evaluated.
For each task, we randomly sample 100 contexts from the test set and obtain text generated under these contexts by different models. For each pair of generated text, three annotators are asked to give a preference (win, lose, or tie) in terms of three metrics respectively. We adopt majority voting to make final decisions among the three annotators.


\noindent\textbf{Implementation Details.}
Our implementation is based on Transformers\footnote{https://github.com/huggingface/transformers}~\citep{wolf2020transformers} and PyTorch~\citep{paszke2019pytorch}. For the GPT2 baseline, we use the GPT2-small pre-trained model~\citep{radford2019language}. Specifically, the dimension of word embedding and the dimension of hidden vectors are set to 768. The number of self-attention blocks is set to 12 and 12 heads are used in self multi-head attention. For the mixed-granularity decoder in our framework, we apply the same model size as GPT2-small. For SSR-AR, we use a structure similar to GPT2 and we use a structure similar to the uncased BERT-base pre-trained model \citep{devlin2019bert} for SSR-NonAR.
We train the model using Adam \citep{kingma2014adam} with learning rate 0.00005. The dropout rate is set to 0.1 for regularization. Following \cite{fan2018hierarchical} we generate text with random top $k$ sampling, where next words are sampled from the top $k=20$ candidates rather than the entire vocabulary distribution. 
We conduct experiments on 1 NVIDIA Tesla V100 GPU.

\section{Results and Discussions}

\subsection{Main Results}

\textbf{Story Ending Prediction}
We verify the performance of ``surface realization via matching'' on the task of story ending prediction.
Automatic evaluation results are shown in Table~\ref{table:story_ending_prediction}. We can see that our models  SSR-AR(II) and SSR-NonAR(II) outperform GPT2(FT) significantly. This shows the effectiveness of our proposed framework. GPT2(FT) employs the perplexity to conduct sentence selection, while SSR-AR(II) and SSR-NonAR(II) are sentence-level matching. That is, the perplexity depends on the probabilities of all tokens in candidate sentences.
In contrast, we first predict the next sentence vector and then compute the similarity of the predicted sentence vector with the candidate sentence vectors. 
The evaluation results prove that sentence-level selection is better than token-level selection.
Compared with SSR-AR(II) and SSR-NonAR(II), the performance of SSR-AR(I) and SSR-NonAR(I) drop a lot, indicating the importance of contrastive loss during training. In addition, SSR-NonAR(II)+PT further improves the performance. This demonstrates that our framework can further improve by pre-training SSR on the extra corpus.

\begin{figure*}[!t]
\centering
\begin{minipage}[t]{0.49\columnwidth}
\centering
\includegraphics[width=1\columnwidth]{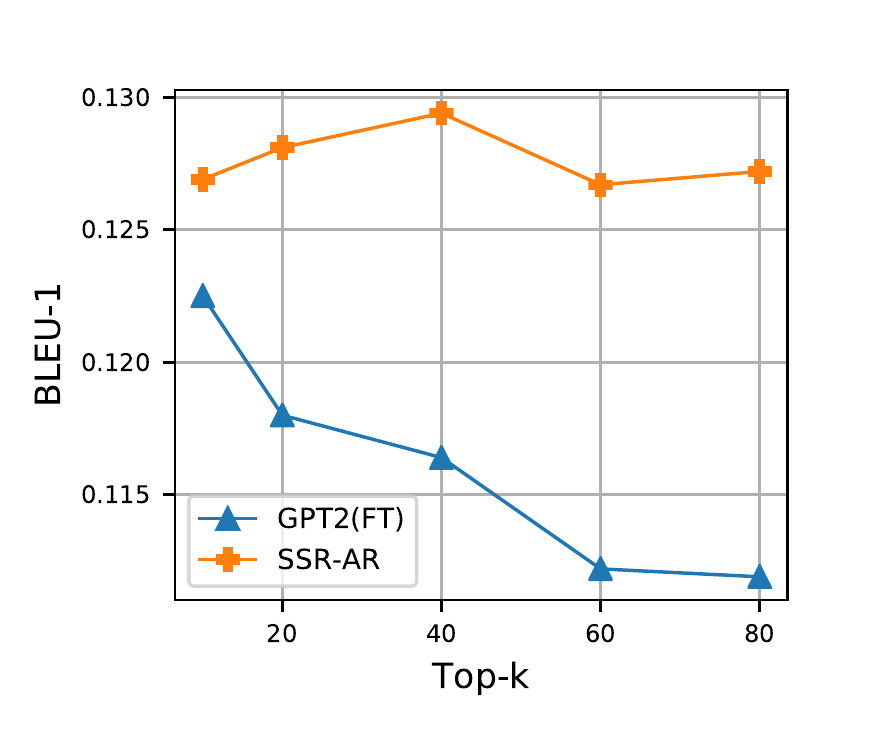}
\subcaption{BLEU-1}
\end{minipage}
\centering
\begin{minipage}[t]{0.49\columnwidth}
\centering
\includegraphics[width=1\columnwidth]{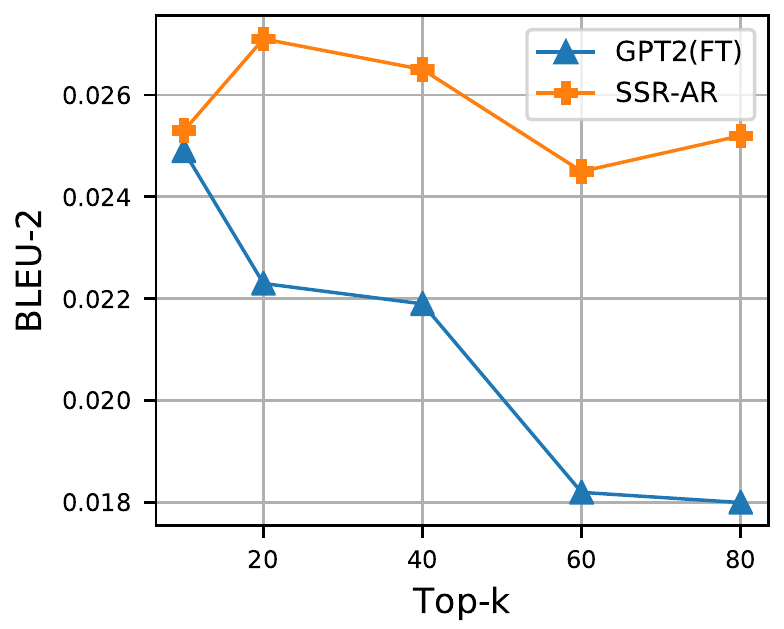}
\subcaption{BLEU-2}
\end{minipage}
\centering
\begin{minipage}[t]{0.49\columnwidth}
\centering
\includegraphics[width=1\columnwidth]{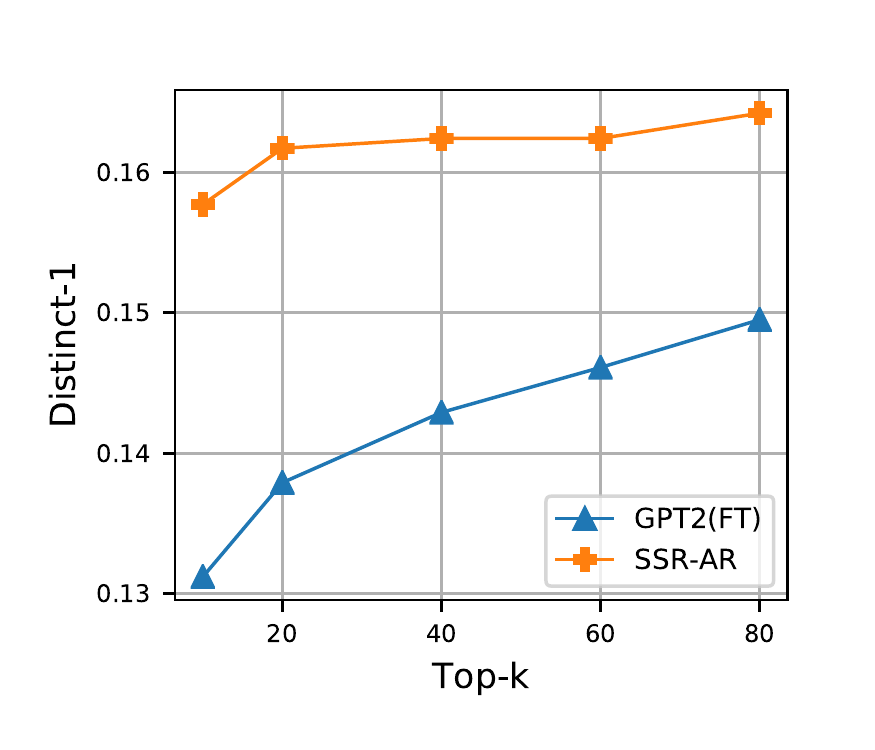}
\subcaption{Distinct-1}
\end{minipage}
\centering
\begin{minipage}[t]{0.49\columnwidth}
\centering
\includegraphics[width=1\columnwidth]{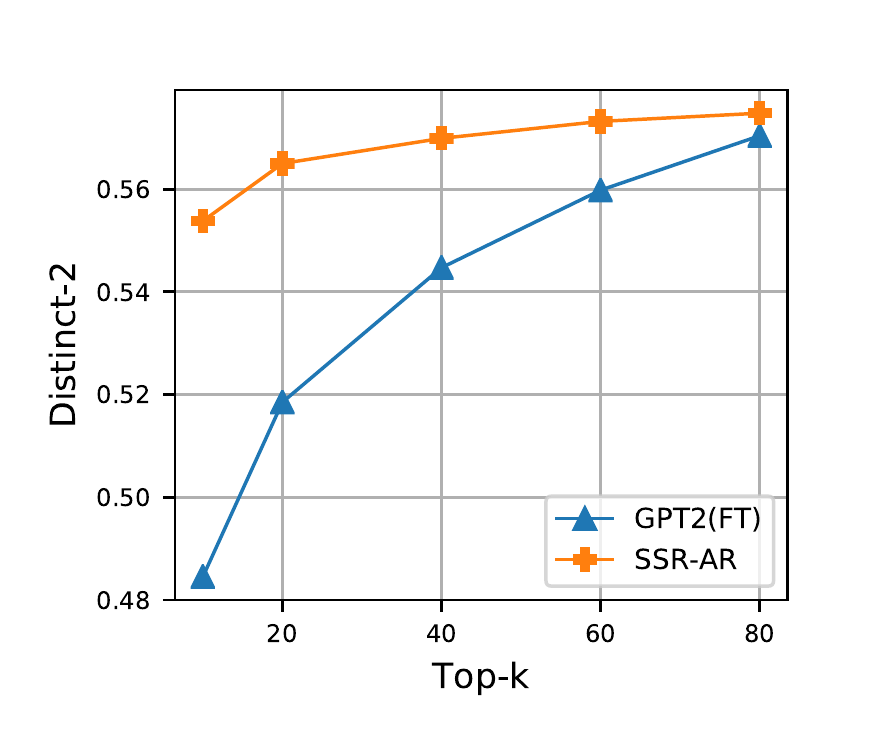}
\subcaption{Distinct-2}
\end{minipage}

\vspace{-1mm}
\caption{Topic stability analysis of SSR and GPT2 according to the hyperparameter $k$ in top-$k$ sampling.}
\vspace{-2mm}
\label{fig:topic_k}
\end{figure*}

\begin{table*}[!t]
\renewcommand\arraystretch{0.8}
\centering
\small
\resizebox{1.8\columnwidth}{!}{
\begin{tabular}{@{}l|cccc|cccc@{}}
\Xhline{3\arrayrulewidth}
                 & \textbf{B-1} &\textbf{B-2} &\textbf{B-3} & \textbf{B-4} & \textbf{D-1} & \textbf{D-2}  & \textbf{D-3}  & \textbf{D-4}  \\
\hline
GPT2(FT)  & 0.0658 & 0.0207 & \underline{0.0049} & \textbf{0.0029} & 0.2347     & 0.6950     & 0.9206     & 0.9808     \\
\hline
SSR-AR(I)  & \textbf{0.0854} & \textbf{0.0239} & 0.0033 & 0.0002 & 0.1608     & 0.5090     & 0.7575     & 0.8877     \\
SSR-AR(II)  & 0.0583 & 0.0177 & 0.0020 & 0.0002 & \underline{0.2583}     & \underline{0.7482}     & \underline{0.9495}     & \underline{0.9891}     \\
SSR-AR     & \underline{0.0693} & \underline{0.0232} & \textbf{0.0054} & \underline{0.0017} & \textbf{0.2565}     & \textbf{0.7597}     & \textbf{0.9549}     & \textbf{0.9914}     \\
\Xhline{3\arrayrulewidth}
\end{tabular}
}
\vspace{-1mm}
\caption{Automatic evaluation results of dialogue generation.}
\label{table:dialogue_generaton}
\vspace{-3mm}
\end{table*}

\begin{table*}[!t]
\renewcommand\arraystretch{0.8}
\centering
\small
\resizebox{1.8\columnwidth}{!}{
\begin{tabular}{@{}l|cccc|cccc@{}}
\Xhline{3\arrayrulewidth}
                 & \textbf{B-1} &\textbf{B-2} &\textbf{B-3} & \textbf{B-4} & \textbf{D-1} & \textbf{D-2}  & \textbf{D-3}  & \textbf{D-4}  \\
\hline
IGPT2(FT) & 0.1191 & 0.0244 & 0.0069 & 0.0021 & 0.1391     & 0.5276     & 0.8480     & 0.9615     \\
\hline
SSR-NonAR(I)         & \textbf{0.1433} & 0.0234 & 0.0070 & 0.0018 & 0.0851     & 0.3412     & 0.6365     & 0.8262     \\
SSR-NonAR(II) & 0.1202 & 0.0246 & 0.0075 & 0.0024 & 0.1615     & 0.5781     & \underline{0.8784}     & \textbf{0.9705}     \\
SSR-NonAR  & 0.1326 & \underline{0.0296} & \underline{0.0105} & \textbf{0.0042} & \underline{0.1669}     & \underline{0.5823}     & \textbf{0.8787}     & \underline{0.9704}     \\
SSR-NonAR+PT        & \underline{0.1413} & \textbf{0.0342} & \textbf{0.0114} & \underline{0.0039} & \textbf{0.1698}     & \textbf{0.5832}     & 0.8725     & 0.9667     \\
 \Xhline{3\arrayrulewidth}
\end{tabular}
}
\vspace{-1mm}
\caption{Automatic evaluation results of sentence infilling.}
\label{table:sentence_infilling}
\vspace{-3mm}
\end{table*}

\begin{table*}[!t]
\renewcommand\arraystretch{0.8}
\centering
\resizebox{2\columnwidth}{!}{
\begin{tabular}{@{}lllllllllll@{}}
 \Xhline{3\arrayrulewidth}
\multirow{2}{*}{\textbf{Task}}   & \multirow{2}{*}{\textbf{Method}}       & \multicolumn{3}{c}{\textbf{Grammaticality}} & \multicolumn{3}{c}{\textbf{Logicality}} & \multicolumn{3}{c}{\textbf{Informativeness}} \\
                        &                               & \textbf{Win}(\%)   & \textbf{Tie}(\%)   & \textbf{Lose}(\%)   &\textbf{ Win}(\%)  & \textbf{Tie}(\%)  & \textbf{Lose}(\%) & \textbf{Win}(\%)    & \textbf{Tie}(\%)   & \textbf{Lose}(\%)   \\ 
\hline
story ending gen. & SSR-AR vs. GPT2(FT)  & 2         & 92        & \textbf{6}          & \textbf{27}       & 57       & 16       & \textbf{11}         & 83        & 6          \\
dialogue gen.     & SSR-AR vs. GPT2(FT)  & 4         & 89        & \textbf{7}          & \textbf{21}       & 61       & 18       & \textbf{9}          & 86        & 5          \\
sentence infi.     & SSR-NonAR vs. IGPT2(FT) & \textbf{8}         & 87        & 5          & \textbf{31}       & 46       & 23       & \textbf{11}         & 81        & 8          \\ 
 \Xhline{3\arrayrulewidth}
\end{tabular}
}
\vspace{-1mm}
\caption{Human evaluation results for three generation tasks.}
\label{table:human_result}
\vspace{-5mm}
\end{table*}

\textbf{Story Ending Generation}
Table \ref{table:story_ending_generaton} reports automatic evaluation results of story ending generation. SSR-AR outperforms GPT2(FT) in most metrics, indicating that the effectiveness of our framework on generation. In addition, SSR-AR(I) performs worse than GPT2(FT) in most metrics while SSR-AR(II) outperforms GPT2(FT) in Distinct. This indicates that the contrastive loss contributes to improving diversity. Without the contrastive loss, SSR-AR(I) tends to generate common words. Thus, it achieves a higher BLEU-1 score and a lower Distinct score than GPT2(FT). Compared with SSR-AR(II), SSR-AR improves BLEU score significantly. This shows that the mixed-granularity decoder is helpful to reproduce words in contexts and thus the results have better relevance to contexts. 
 
\textbf{Dialogue Generation}
The trend in Table \ref{table:dialogue_generaton} is similar to that in Table \ref{table:story_ending_generaton}. SSR-AR performs better than GPT2(FT) in most metrics, and the contrastive loss and the mixed-granularity sentence decoder improve the performance of Distinct and BLEU respectively. 

\textbf{Sentence Infilling}
Automatic evaluation results of sentence infilling are shown in Table~\ref{table:sentence_infilling}. We can see that SSR-NonAR outperforms IGPT2(FT) in all metrics. Similar to SSR-AR(I), SSR-NonAR(I) also performs worse in Distinct. When adding the contrastive loss, SSR-NonAR(II) performs better than IGP2(FT). This demonstrates that the contrastive loss is important for both SSR-AR and SSR-NonAR. Compared with SSR-NonAR(II), SSR-NonAR improves BLEU further, indicating the effectiveness of the mixed-granularity decoder. 
In addition, when pre-training the sentence-level language model on the extra corpus, SSR-NonAR+PT outperforms SSR-NonAR in most metrics, showing that our framework can benefit from pre-training. 
From the results of three generation tasks, we can see that the contrastive loss and the mixed-granularity decoder contribute to Distinct and BLEU respectively. With these two components, SSR obtains better performance than the token-level method.

\textbf{Human Evaluation}
For three generation tasks, we further utilize human evaluation to evaluate our framework. The results are shown in Table \ref{table:human_result}. SSR outperforms GPT2(FT) in terms of Logicality and Informativeness.
Considering that we first predict a sentence vector and then decode it into a sentence, the text generated by this process is much more consistent than that produced via a token by token method. We will prove this in a more in-depth analysis in the next section. In addition, SSR performs close to GPT2(FT) in terms of Grammaticality. We claim that both methods use an AR-based decoder, so the fluency of generated sentences is close.

\subsection{Analysis of Topic Drift}

In order to investigate the degree of topic drift of different models, we conduct generation with different top-k settings under the same contexts on story ending generation. From Figure~\ref{fig:topic_k} we can observe that BLEU of GPT2(FT) drops sharply and Distinct of GPT2(FT) increases a lot when increasing $k$, showing that the larger $k$ is, the more serious the topic drift is.
While BLEU of our method drops a little and maintains a stable range. This indicates that SSR can control the semantic of the content to be generated through the predicted sentence vector and effectively avoid topic drift. 

Table \ref{name_example} depicts generated results containing named entities. With the mixed-granularity decoder, SSR-AR can restore named entities accurately, 
indicating that the mixed-granularity decoder contributes to better consistency of the surface realization. 
Table \ref{surface_example} shows examples generated by top-k sampling four times. The semantics of the generated results by GPT2(FT) vary greatly and are not consistent with the context. The topic drift phenomenon occurs. In contrast, SSR-AR(II) predicts context-consistent topics, generating sentences about ``how to build''. But the fine-grained words are quite different from the ones in the context. Jointly considering the predicted sentence vector and words from the context, SSR-AR produces sentences that are more consistent and relevant with the context.

\begin{table}[!t]
\renewcommand\arraystretch{0.5}
\centering
\begin{tabular}{ll} 
\toprule

\multicolumn{2}{p{0.95\columnwidth}}{\small \textbf{Context:} \textbf{Rex} had always wanted to play hockey. He decided that he first needed to learn to skate. He practiced skating every day until he was an expert. Finally \textbf{Rex}  was able to begin playing hockey.			}                                                             \\
\midrule
\small \textbf{GPT2(FT)}
 & \multicolumn{1}{p{0.7\columnwidth}}{ \small \textbf{1:}    \textbf{Rex} is now a very good skater.}
                         \\
                      \midrule
\small \textbf{SSR-AR(II)}
& \multicolumn{1}{p{0.7\columnwidth}}{\small \textbf{1:}  \textbf{Andrew} is very proud to now experience his playing ability! }
      \\
                      \midrule
\small \textbf{SSR-AR}
& \multicolumn{1}{p{0.7\columnwidth}}{\small \textbf{1:}  \textbf{Rex} is very proud that he now has great hockey skills!  }
                              \\
                      \bottomrule
\end{tabular}
\vspace{-1mm}
    \caption{Generated examples containing named entities.}
	\label{name_example}
\vspace{-3mm}
\end{table}

\begin{table}[!t]
\renewcommand\arraystretch{0.5}
\centering
\begin{tabular}{ll} 
\toprule
\multicolumn{2}{p{0.95\columnwidth}}{\small \textbf{Context:} Willie had too much stuff. Willie bought a shed to store all his stuff. Willie had a hard time putting up the shed. He called some friends for help.			}                                                             \\
\midrule
\multirow{4}{*}{\small \textbf{GPT2(FT)}} 
 & \multicolumn{1}{p{0.7\columnwidth}}{\small \textbf{1:}   Willie sold his shed and made enough money to pay for the house.   }                         \\
                       & \multicolumn{1}{p{0.7\columnwidth}}{\small \textbf{2:}  After a few days they bought the shed. }                             \\
                       & \multicolumn{1}{p{0.7\columnwidth}}{\small \textbf{3:}   They brought him the shed and fixed the shed.    }                        \\
                       & \multicolumn{1}{p{0.7\columnwidth}}{\small \textbf{4:}   The shed became a nice storage place. }
                              \\
                      \midrule
\multirow{2}{*}{\small \textbf{SSR-AR(II)}} 
 & \multicolumn{1}{p{0.7\columnwidth}}{\small \textbf{1:}  They made it all back in the tools.  }                          \\
                       & \multicolumn{1}{p{0.7\columnwidth}}{\small \textbf{2:} They helped keep the pile together plenty.             }                 \\
                       & \multicolumn{1}{p{0.7\columnwidth}}{\small \textbf{3:} They mopped and hauled the house out. }
                              \\
                       & \multicolumn{1}{p{0.7\columnwidth}}{\small \textbf{4:} They finished the shed and put everything back. }
                              \\
                      \midrule
\multirow{4}{*}{\small \textbf{SSR-AR}}   
& \multicolumn{1}{p{0.7\columnwidth}}{\small \textbf{1:}  They helped put up the shed floor.
  }                               \\
                      & \multicolumn{1}{p{0.7\columnwidth}}{\small \textbf{2:} They helped pile the shed down.
}                 \\
                      & \multicolumn{1}{p{0.7\columnwidth}}{\small \textbf{3:}   They helped put the shed out and usable.
      }                         \\
                      & \multicolumn{1}{p{0.7\columnwidth}}{\small \textbf{4:}  They helped haul the shed all out.
        }                      \\
                      \bottomrule
\end{tabular}
\vspace{-1mm}
    \caption{Generated results by top-$k$ for 4 rounds.}
	\label{surface_example}
	\vspace{-5mm}
\end{table}

\vspace{-3mm}
\section{Conclusion}\label{sec:conclusion}
We propose a new framework named Sentence Semantic Regression (SSR) for text generation. Two architectures SSR-AR and SSR-NonAR are designed to conduct sentence semantic regression autoregressively and bidirectionally. A mixed-granularity decoder is used to generate sentences considering the predicted sentence vectors and the context tokens.
We conduct experiments on the tasks of story ending prediction, story ending generation, dialogue generation, and sentence infilling. The results show SSR obtain better performance in terms of automatic metrics and human evaluation.



\bibliographystyle{acl_natbib}
\bibliography{anthology,acl2021}


\end{document}